%% file: main.tex
\pdfoutput=1
\documentclass{article}
\PassOptionsToPackage{numbers, compress}{natbib}

\usepackage{caption}        %
\usepackage[utf8]{inputenc} %
\usepackage[T1]{fontenc}    %
\usepackage{hyperref}       %
\usepackage{url}            %
\usepackage{booktabs}       %
\usepackage{amsfonts}       %
\usepackage{nicefrac}       %
\usepackage{microtype}      %
\usepackage{xcolor}         %
\input{preamble}

\usepackage[main, final]{neurips_2026}

\title{MatrixSplat: Matrix-Parallel and Feature-Scalable 3D Gaussian Splatting Rendering on GPUs}

\author{%
  Yangming Zhang$^{1}$ \quad Siyi Wu$^{1}$ \quad Jian Wang$^{2}$ \quad Bingzhe Li$^{3}$ \\
  \textbf{Wei Niu}$^{4}$ \quad \textbf{Sheng Di}$^{5}$ \quad \textbf{Ming Li}$^{1}$ \quad \textbf{Miao Yin}$^{1}$ \\
  $^{1}$The University of Texas at Arlington \quad $^{2}$Snap Inc. \quad $^{3}$The University of Texas at Dallas \\
  $^{4}$University of Georgia \quad $^{5}$Argonne National Laboratory \\
  \small\texttt{\{yxz0925, sxw8121\}@mavs.uta.edu} \quad \texttt{jwang4@snapchat.com} \quad \texttt{bingzhe.li@utdallas.edu} \\
  \small\texttt{wniu@uga.edu} \quad \texttt{sdi1@anl.gov} \quad \texttt{\{ming.li, miao.yin\}@uta.edu} \\
}

\begin{document}

\maketitle

\begin{abstract}

3D Gaussian Splatting (3DGS) has emerged as a mainstream representation for real-time novel view synthesis and 3D reconstruction. However, existing 3DGS rendering pipelines are poorly aligned with modern GPU architectures: depth-sorted alpha compositing introduces strong per-pixel sequential dependencies, limiting parallelism and resulting in scalar Fused Multiply-Add (FMA)-dominated execution that underutilizes high-throughput matrix units such as Tensor Cores. This mismatch becomes more severe for multimodal and semantically enriched 3DGS, where high-dimensional features must be accumulated from millions of Gaussians, leading to rapidly increasing computation and memory costs. To address these challenges, we propose MatrixSplat, a matrix-parallel rendering framework for feature-scalable 3DGS on modern GPUs. We first identify the sequential dependency of alpha compositing as a key obstacle to efficient high-dimensional feature rendering, and adopt an order-independent weighted accumulation formulation that transforms feature rendering into matrix-oriented computation. Our framework features three key designs: \textit{i)} a tile-based splatting pipeline that maps feature accumulation to warp-level matrix-multiply-accumulate execution on Tensor Cores; \textit{ii)} a warp-granularity Gaussian packing strategy that mitigates visibility irregularity and improves feature reuse; and \textit{iii)} a lazy channel-batched feature-fetching pipeline that reduces shared-memory pressure, maintains high warp occupancy, and overlaps memory access with matrix execution for high-dimensional features.
Experiments on Mip-NeRF 360, LERF, and 3D-OVS demonstrate both high performance and strong scalability for feature-rich 3DGS rendering. Particularly, our method achieves up to 3.8$\times$ speedup over 3DGS and 2$\times$ over TC-GS while maintaining rendering quality and semantic accuracy, and further reduces render-kernel time by up to 36.7$\times$ for high-dimensional features. Integrated into existing fast renderers, it provides an additional 1.69--2.84$\times$ speedup.
\end{abstract}

\section{Introduction}
\label{sec:introduction}

\begin{figure}[t]
  \centering
  \vspace{-1mm}
  \begin{minipage}[t]{0.48\textwidth}
    \centering
    \includegraphics[width=\linewidth]{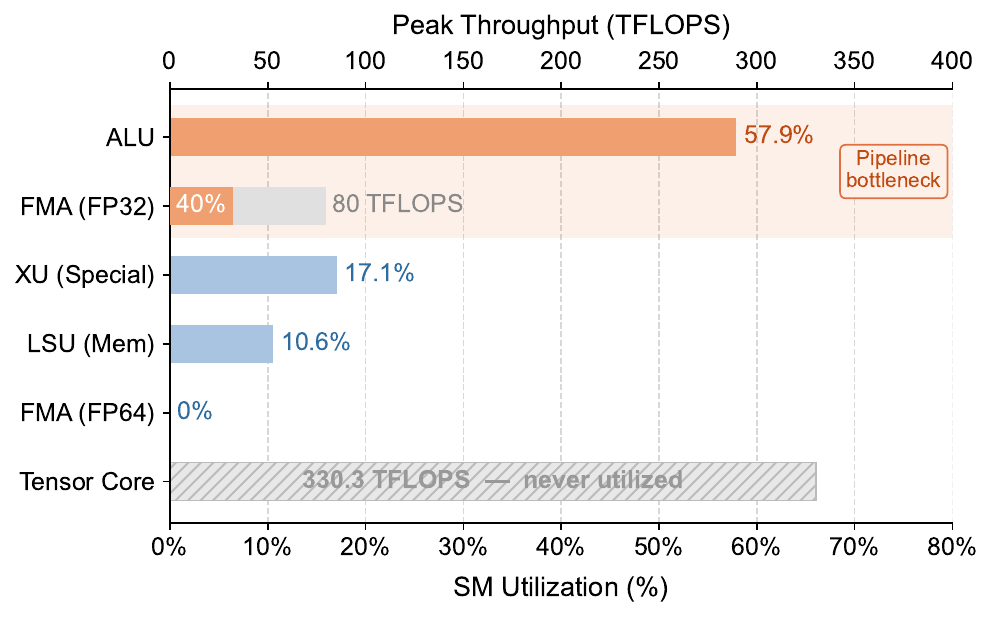}
    \vspace{-7mm}
    \caption{\textbf{SM utilization of 3DGS rendering.} Rendering is dominated by ALU and FMA (FP32), leaving Tensor Core never utilized.} %
    \label{fig:sm-utilization}
  \end{minipage}
  \hspace{1mm}
  \begin{minipage}[t]{0.48\textwidth}
    \centering
    \includegraphics[width=\linewidth,height=4cm,keepaspectratio]{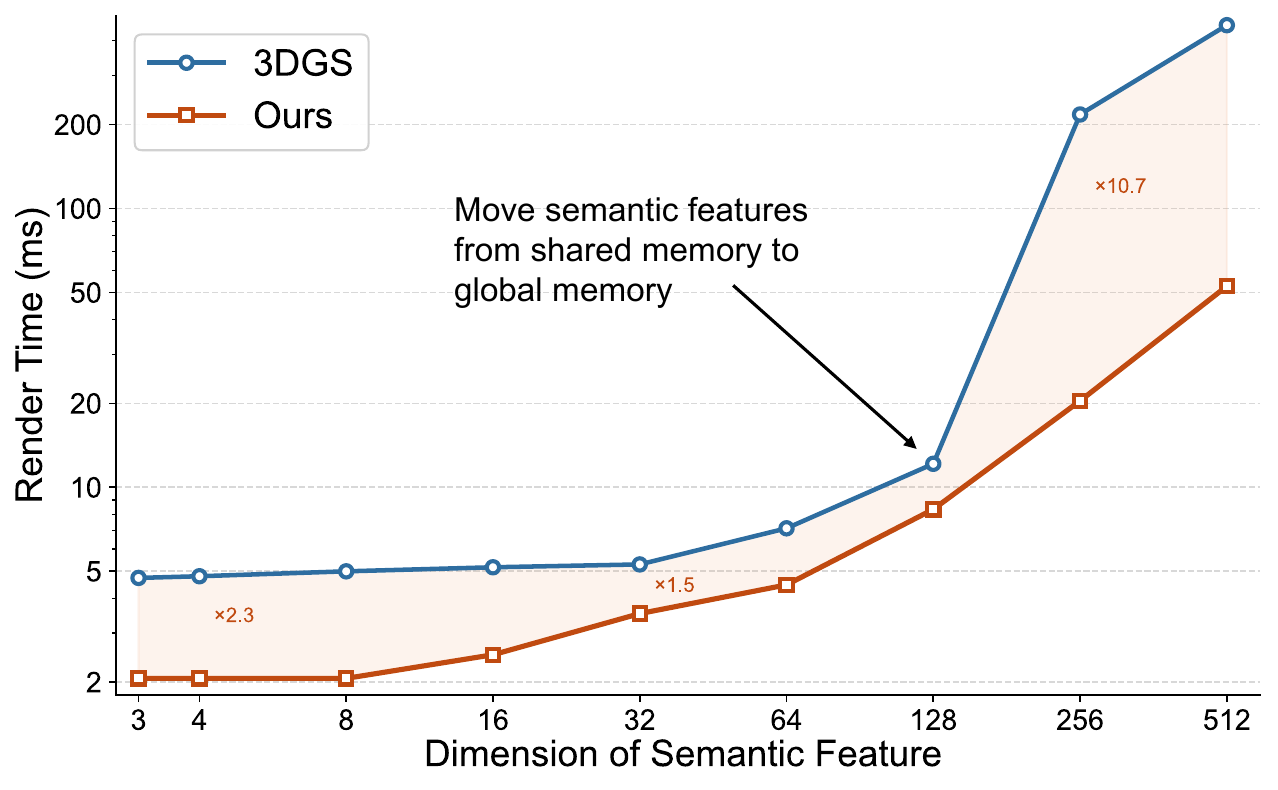}
    \vspace{-3mm}
    \caption{\textbf{Semantic feature scalability.} Rendering performance degrades dramatically with increasing feature dimensionality.}%
    \label{fig:sweep-plot}
  \end{minipage}
  \vspace{-2mm}
\end{figure}

Modern GPU architectures are increasingly optimized for high-throughput matrix computation, yet many emerging graphics and vision workloads fail to fully exploit these capabilities. 3D Gaussian Splatting (3DGS)~\cite{kerbl20233d} has recently emerged as a mainstream paradigm for novel view synthesis and 3D reconstruction~\cite{chen2024survey,hanson2025speedy,jia2026splaxel}. It represents a scene as a set of 3D Gaussian primitives, each parameterized by spatial position, scale, opacity, rotation, and spherical harmonic (SH) coefficients, enabling differentiable rendering with high visual quality and real-time performance~\cite{bagdasarian20243dgs}. However, the standard 3DGS pipeline relies on depth-sorted alpha compositing, where Gaussian contributions are accumulated in a strict front-to-back order. This design introduces strong per-pixel sequential dependencies, limiting parallelism and causing frequent instruction stalls, with execution dominated by scalar Fused Multiply-Add (FMA) operations~\cite{liao2025tc,feng2024flashgs,du2026mobile}. Consequently, the rendering pipeline is poorly aligned with modern matrix-computation-oriented GPU architectures, leaving high-throughput matrix units largely underutilized. As shown in Fig.~\ref{fig:sm-utilization}, profiling of Streaming Multiprocessor (SM) utilization reveals that rendering is dominated by Arithmetic Logic Unit (ALU) and FP32 FMA operations, while Tensor Cores remain entirely unused despite offering up to 4$\times$ higher peak throughput. This mismatch exposes a \textbf{\textit{fundamental gap}} between the dependency structure of current 3DGS rendering algorithms and the computing patterns of modern GPUs.

This challenge becomes significantly more severe as 3DGS evolves toward multimodal and semantically enriched representations. Recent advances extend 3DGS beyond photorealistic reconstruction by integrating high-dimensional features, enabling support for emerging applications such as world models in robotics, embodied AI, and autonomous driving~\cite{liu2026palm,shen2025fine}. For example, prior works~\cite{qin2024langsplat,li2025semanticsplat,jiang2024vr} embed high-dimensional language features into Gaussian primitives, enabling semantically aligned 3D understanding without relying on expensive and often low-quality 3D annotations. However, existing 3DGS rendering pipelines are not designed for such \textbf{\textit{feature-scalable}} workloads, and the algorithm-hardware mismatch is therefore further magnified. As feature dimensionality increases, the accumulation workload over millions of Gaussians grows rapidly, resulting in substantially higher memory traffic and computational cost~\cite{huang2025seele}. As shown in Fig.~\ref{fig:sweep-plot}, the original 3DGS can no longer stage the full feature vector in shared memory when the dimension is greater than 128, so features are fetched from global memory. Rendering performance degrades dramatically with increasing feature dimensionality~\cite{li2025semanticsplat,zhou2024feature}, especially once features exceed the shared-memory capacity.

Prior works~\cite{deng2024efficient,fan2023lightgaussian,mallick2024taming,lee2024compact,zhang2024gaussianspa,ou2026flicker} have attempted to address this mismatch from both algorithmic and hardware perspectives, but they accelerate different stages of the pipeline. Pipeline-level accelerators such as FlashGS~\cite{feng2024flashgs}, Speedy-Splat~\cite{hanson2025speedy}, and AdR-Gaussian~\cite{wang2024adrgaussian} reduce the work \textit{around} blending through tighter culling, scheduling, and sorting, yet still accumulate each pixel with a serial scalar loop. Sorting-free methods~\cite{hou2024sort,kheradmand2025stochasticsplats,liu2026duplex} replace sorted alpha compositing with weighted-sum or stochastic rendering to remove depth sorting, but execute the resulting accumulation with the same scalar FMAs. Tensor-Core approaches such as TC-GS~\cite{liao2025tc} and GEMM-GS~\cite{li2026gemmgs} map the alpha computation onto matrix units, while the blending loop itself remains order-dependent. Therefore, a gap remains: no existing method uses order independence to express the feature accumulation itself as matrix multiplication, even though this is the part whose cost grows with feature dimensionality~\cite{ha2026langgs}.

To bridge this algorithm-hardware gap, we propose MatrixSplat, a matrix-parallel framework for feature-scalable 3DGS rendering on modern GPUs. Our key insight is that the sequential dependency introduced by alpha compositing is what prevents feature accumulation from being executed as matrix operations. By revisiting the rendering pipeline, we derive an order-independent weighted accumulation formulation that removes this dependency and transforms scalable feature rendering into matrix-oriented computation. Building on this formulation, we design a scalable tile-based splatting pipeline that maps feature accumulation to warp-level packed matrix-multiply-accumulate (MMA) execution. To further support high-dimensional semantic features, we introduce a lazy channel-batched feature-fetching strategy that preserves data reuse, reduces shared-memory pressure, and maintains high execution efficiency as feature dimensionality grows. Together, these designs enable efficient and scalable 3DGS rendering by leveraging high-throughput matrix units on modern GPUs. Because our method changes how blending is executed rather than how much work reaches it, it is complementary to pipeline-level accelerators and can be combined with them.

Our contributions are summarized as follows:
\begin{itemize}

\item We analyze the 3DGS rendering pipeline and identify the sequential dependency of alpha compositing as a key bottleneck for feature-embedded Gaussian rendering on modern GPUs. Inspired by prior order-independent rendering ideas, we tailor weighted accumulation to semantic feature rendering and reorganize the resulting dependency-free computation into a hardware-aligned, matrix-oriented execution pattern.

\item We propose a warp-level, matrix-oriented design that enables efficient mapping of feature-scalable rendering to Tensor Core MMA execution. By packing valid Gaussians at warp level, our method enables efficient matrix-computation-friendly feature accumulation while reducing visibility irregularity. This design improves the reuse of Gaussian features.

\item We develop a scalable, lazy, channel-batched feature-fetching pipeline. This lazy loading reduces shared-memory pressure as feature dimensionality grows, preserves higher active warps per SM, and hides latency by overlapping memory loading and MMA execution, thereby supporting efficient rendering of high-dimensional features.

\end{itemize}
We evaluate our method on Mip-NeRF 360, LERF, and 3D-OVS. Experiments show that our method achieves up to 3.8$\times$ end-to-end speedup over 3DGS and 2$\times$ over TC-GS while maintaining visual quality and semantic accuracy, and further reduces render-kernel time by up to 36.7$\times$ for high-dimensional semantic features. It is also 2.2--3.5$\times$ faster than Sort-Free rendering and provides an additional 1.69--2.84$\times$ speedup when integrated into existing fast renderers.

\section{Related Works}
\label{Related_Works}
\begin{wraptable}{r}{0.5\linewidth}
  \centering
  \vspace{-4.5mm}
  \caption{\textbf{Positioning of related methods.} TC: Tensor Cores.}
  \label{tab:related}
  \vspace{-2mm}
  \resizebox{\linewidth}{!}{%
  \begin{tabular}{llll}
    \toprule
    \textbf{Method} & \textbf{Stage} & \textbf{Order} & \textbf{Dim.} \\
    \midrule
    FlashGS, Speedy-Splat, & \multirow{2}{*}{Culling, scheduling} & \multirow{2}{*}{Sorted} & \multirow{2}{*}{RGB} \\
    AdR-Gaussian, HiGS & & & \\
    TC-GS, GEMM-GS & Alpha (TC) & Sorted & RGB \\
    ORANGE & Blending (NPU) & Sorted & RGB \\
    Sort-Free, StochasticSplats & Sorting & None & RGB \\
    Duplex-GS & Sorting & Proxy & RGB \\
    Feature-3DGS, LangSplat(V2) & Decoding & Sorted & Compr. \\
    \midrule
    \textbf{MatrixSplat} & \textbf{Accumulation (TC)} & \textbf{None} & \textbf{Native} \\
    \bottomrule
  \end{tabular}%
  }
  \vspace{-3mm}
\end{wraptable}

\paragraph{Feature-rich 3D Gaussian Splatting.}
Gaussian Grouping~\cite{ye2024gaussian}, LangSplat~\cite{qin2024langsplat}, and Semantic Gaussians~\cite{guo2026semantic} attach identity or CLIP-aligned features to Gaussians for segmentation, editing, and open-vocabulary queries. Because rendering cost grows with feature dimensionality, these systems usually compress features before rendering: Feature-3DGS~\cite{zhou2024feature} renders a low-dimensional field followed by a decoder, LangSplat compresses 512-D CLIP features into a 3-D latent space, and LangSplatV2~\cite{li2025langsplatv2} replaces decoding with sparse coefficient splatting. ClunoGS~\cite{xu2026clunogs} further compacts language Gaussians by jointly sparsifying and quantizing them in a unified optimization. Since feature accumulation remains a dominant cost~\cite{ye2025gaussian}, we accelerate it directly so that features can be rendered at their native dimension.

\paragraph{Fast 3DGS rendering.}
Pipeline-level accelerators reduce the work around blending while keeping in-order compositing (Tab.~\ref{tab:related}). StopThePop~\cite{radl2024stopthepop} adds hierarchical re-sorting and culling for view-consistent rendering, FlashGS~\cite{feng2024flashgs} combines precise Gaussian-tile intersection with kernel-level scheduling, Speedy-Splat~\cite{hanson2025speedy} tightens Gaussian localization and prunes primitives, AdR-Gaussian~\cite{wang2024adrgaussian} culls with adaptive radii and balances load across threads, and HiGS~\cite{pajak2026higs} decouples binning and rasterization tile sizes. Compression and pruning methods~\cite{fan2023lightgaussian,lee2024compact,mallick2024taming,zhang2024gaussianspa} reduce the number of Gaussians, and dedicated accelerators~\cite{ou2026flicker,ye2025gaussian} design hardware for 3DGS rendering. Matrix-unit methods such as TC-GS~\cite{liao2025tc}, GEMM-GS~\cite{li2026gemmgs}, and ORANGE~\cite{li2026orange} map alpha computation or blending onto Tensor Cores or NPU matrix units, but their blending loop remains order-dependent. We instead matrixize the accumulation itself, so our method is complementary to these approaches and can be combined with them (Sec.~\ref{sec:comparison}).

\paragraph{Order-independent rendering.}
Weighted blended OIT~\cite{everitt2001interactive,mcguire2013weighted} replaces ordered compositing with commutative weighted accumulation, which avoids sorting and the popping artifacts caused by unstable depth order. In 3DGS, Sort-Free Gaussian Splatting~\cite{hou2024sort} learns depth-dependent weights through linear-correction weighted-sum rendering, StochasticSplats~\cite{kheradmand2025stochasticsplats} replaces sorting with stochastic rasterization, and Duplex-GS~\cite{liu2026duplex} combines proxy-level ordering with weighted blending to improve quality. These methods use order independence to avoid sorting, but still accumulate each pixel with scalar FMAs. We instead use it to change the compute primitive, turning the per-pixel accumulation into dense matrix products on Tensor Cores.

\section{Background}
\label{BACKGROUND}
\noindent\textbf{Order-Independent Transparency. }
Rendering semi-transparent scenes requires modeling partial coverage, which is commonly achieved with the \textit{over} operator~\cite{porter1984compositing}. Like Eq.~\ref{eqn:render-color}, it is inherently order-dependent, because the transmittance of each primitive depends on all preceding primitives, introducing a loop-carried dependency during blending.

To mitigate the cost of sorting, many approximate compositing methods have been proposed. A notable advancement is weighted blended order-independent transparency (OIT)~\cite{mcguire2013weighted}. This approach replaces conventional front-to-back alpha blending with a commutative weighted-sum accumulation:
\begin{equation}\label{eqn:wieghted-sum}
\boldsymbol{C}=\frac{\sum_{i\in \mathcal{N}} \boldsymbol{c}_i \cdot w\left(\alpha_i, d_i\right)}{\sum_{i\in \mathcal{N}} w\left(\alpha_i, d_i\right)}\left(1-\prod_{i\in \mathcal{N}}\left(1-\alpha_i\right)\right)+\boldsymbol{c}_0 \prod_{i\in \mathcal{N}}\left(1-\alpha_i\right),
\end{equation}
where $\boldsymbol{c}_0$ is the background color, and $\boldsymbol{c}_i$ and $\alpha_i$ represent the color and opacity of the $i$-th primitive. The function $w\left(\alpha_i, d_i\right)$ denotes a weight that decreases with distance. This approach eliminates the sorting stage and avoids visual ``popping'' artifacts caused by unstable depth ordering.

Based on this idea, Linear Correction Weighted Sum Rendering (LC-WSR)~\cite{dai2017deformable,hou2024sort} further refines occlusion modeling by assigning depth-dependent weights through a learnable linear function:
\begin{equation}\label{eqn:LC-WSR}
w(\alpha_i,d_i)
=
\alpha_i \max\left(0, 1-\frac{d_i}{\sigma}\right) v_i,
\quad i=1,2,\cdots,N.
\end{equation}
where $\sigma$, $v_i$ are learnable parameters and $\alpha_i$ is the projected opacity of the $i$-th Gaussian at the pixel. LC-WSR assigns larger weights to near Gaussians while suppressing distant ones to zero.

\paragraph{3D Gaussian Splatting.}
Standard 3D Gaussian Splatting (3DGS)~\cite{kerbl20233d} represents a scene as a set of optimized anisotropic 3D Gaussians. Each Gaussian $g_i$ is defined by its center position $\boldsymbol{\mu}_i \in \mathbb{R}^3$, covariance matrix $\boldsymbol{\Sigma}_i$, opacity $o_i$, and spherical harmonics (SH) coefficients for view-dependent color. Given an arbitrary 3D position $\boldsymbol{x}$, the opacity contribution of the $i$-th Gaussian is defined as
\begin{equation}\label{eqn:opacity-contribution}
\alpha_i(\boldsymbol{x})
=
o_i  \exp\left(
-\frac{1}{2}
(\boldsymbol{x}-\boldsymbol{\mu}_i)^\top
\boldsymbol{\Sigma}_i^{-1}
(\boldsymbol{x}-\boldsymbol{\mu}_i)
\right),
\quad i=1,2,\cdots,N.
\end{equation}

During rendering, each Gaussian is projected onto the image plane with 2D covariance $\boldsymbol{\Sigma}'_i=\boldsymbol{J}_i \boldsymbol{T}_i \boldsymbol{\Sigma}_i \boldsymbol{T}_i^\top \boldsymbol{J}_i^\top$, where $\boldsymbol{T}_i$ is the view transformation and $\boldsymbol{J}_i$ the Jacobian of the local affine approximation of the projection.
For each pixel, the contributing Gaussians are sorted in depth order and composited using alpha blending. The final pixel color $\boldsymbol{C}$ is computed as
\begin{equation}\label{eqn:render-color}
\boldsymbol{C}
=
\sum_{i \in \mathcal{N}}
\boldsymbol{c}_i \alpha_i
\prod_{j=1}^{i-1}
(1-\alpha_j),
\end{equation}
where $\boldsymbol{c}_i$ is the view-dependent color decoded from the SH coefficients, $\alpha_i$ is the projected opacity of the $i$-th Gaussian at the pixel, and $\mathcal{N}$ denotes the depth-ordered set of Gaussians contributing to pixel. %

\paragraph{Language 3D Gaussian Splatting.} LangSplat~\cite{qin2024langsplat} distills 2D region-level CLIP embeddings into 3D Gaussian features, using SAM~\cite{kirillov2023segment} to obtain hierarchical segmentation maps at the subpart, part, and whole levels, denoted $s$, $p$, and $w$. %
Then, LangSplat~\cite{qin2024langsplat} learns a scene-wise autoencoder, where an encoder $E$ compresses the CLIP feature into a low-dimensional latent feature $H_t^l(v)=E\left(L_t^l(v)\right)$, and a decoder reconstructs the original CLIP embedding. Each Gaussian carries hierarchical latent features, rendered with the standard front-to-back alpha compositing:
\begin{equation}\label{eqn:render-feature}
\boldsymbol{F}^l(v)=\sum_{i \in \mathcal{N}} \boldsymbol{f}_i^l \alpha_i \prod_{j=1}^{i-1}\left(1-\alpha_j\right), l \in\{s, p, w\},
\end{equation}
where $\boldsymbol{F}^l(v)$ denotes the rendered latent semantic feature at pixel $v$.
\section{Methodology}
\begin{figure}[t]
    \centering
    \includegraphics[width=\linewidth]{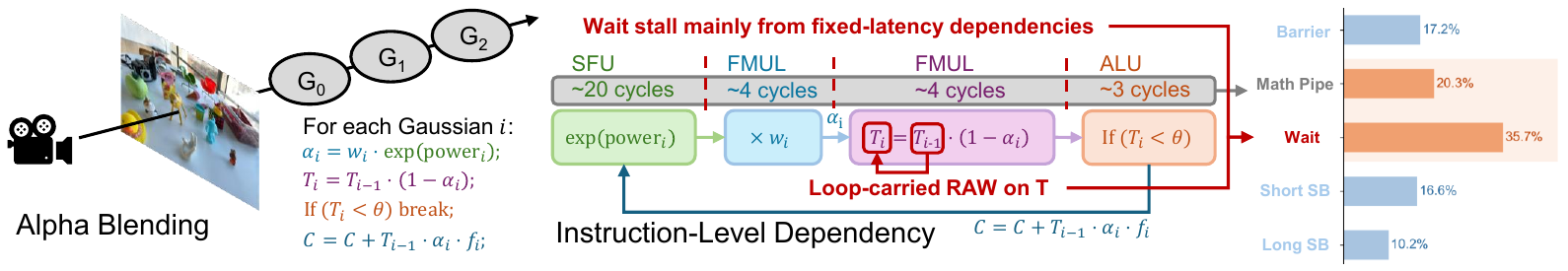}
    \vspace{-7mm}
    \caption{\textbf{Instruction-level dependency and wait stalls.} Conventional front-to-back alpha compositing introduces a long serial instruction sequence chain, leading to a strong wait stall.}%
    \label{fig:stall-breakdown}

  \vspace{-2mm}
\end{figure}

\subsection{Revisiting the Bottlenecks in Feature-Embedded 3DGS Rendering on GPUs}
\label{sec:revisiting}

To analyze the bottlenecks of feature-embedded Gaussian rendering, we first profile the original LangSplat pipeline. As shown in Tab.~\ref{tab:langsplat-latency}, rendering is the dominant stage, accounting for 77\% of the total pipeline latency, and the majority of time is spent in the RenderCUDA kernel. This indicates that the RenderCUDA kernel is the primary system bottleneck.

\begin{wraptable}{r}{0.4\linewidth}
  \centering
  \vspace{-4.5mm}
    \caption{Pipeline latency breakdown.}
    \vspace{-2mm}
  \label{tab:langsplat-latency}
  \resizebox{\linewidth}{!}{
  \begin{tabular}{llr}
    \toprule
    \textbf{Stage} & \textbf{Kernel} & \textbf{\%} \\
    \midrule
    \multirow{4}{*}{Rendering}
      & Preprocessing     &  7.2 \\
      & DuplicateWithKeys &  4.1 \\
      & Sorting           & 14.3 \\ %
      & RenderCUDA        & 51.3 \\
    \midrule
    Decoding        & --- & 19.2 \\
    Post-Processing & --- &  3.8 \\
    \bottomrule
  \end{tabular}
  }
\end{wraptable}
We further analyze the stall behavior of the RenderCUDA kernel in Fig.~\ref{fig:stall-breakdown}. For each Gaussian, the current transmittance update must read back the accumulated result from the previous Gaussians ($T_{j+1} = T_j \cdot (1-\alpha_j)$), forming a loop-carried read-after-write (RAW) dependency that enforces strict sequential execution. The stall breakdown results on the right of Fig.~\ref{fig:stall-breakdown} are consistent with this dependency-limited behavior. The wait (fixed-latency data dependency on arithmetic operations) is the dominant stall type, primarily the computation instruction, with approximately 32 cycles of fixed latency. While the short scoreboard (short SB)  accounts for only 10.2\%. Critically, for the profiled workloads, these results suggest that the kernel is dependency-limited to a significant degree, rather than purely memory-bound. This loop-carried transmittance dependency prevents any cross-iteration instruction overlap within a warp, limiting the modern GPU's ability to hide instruction latency. %

Inspired by those observations, we adopt order-independent blending to a parallel-friendly pattern and further reorganize high-dimensional feature rendering into a hardware-aligned pipeline.

\subsection{Order-Independent Embedded Feature Blending}
To remove the sequential dependency of alpha compositing (where each Gaussian’s contribution depends on the accumulated transmittance of preceding Gaussians), we reformulate semantic Gaussian rendering as an order-independent weighted accumulation over Gaussian embeddings.%

As shown in Eq.~\ref{eqn:render-color} and~\ref{eqn:render-feature}, the rendering process for RGB color and semantic features of Gaussian $g$ is essentially the same. Depth can also be included as an additional rendering attribute. Therefore, we concatenate these attributes into a unified embedded feature $\boldsymbol{e}_i$,
\begin{equation}\label{eqn:concat-feature}
\boldsymbol{e}_i^{\top} =
\operatorname{concat}
\left(
\boldsymbol{c}_i,\ \boldsymbol{f}_i^l,\ [d_i]
\right),
\quad
\boldsymbol{e}_i \in \mathbb{R}^{3 + C + [1]},
\end{equation}
where $[d_i]$ denotes an optional depth and $C$ is the dimension of $\boldsymbol{f}_i^l$ mentioned in Eq.~\ref{eqn:render-feature}, so the embedding dimension is $D = 3 + C\,[+1]$. We assign each valid Gaussian $g$ a pixel-dependent contribution weight $w_{p,g}$ and compute the normalized per-pixel embedding $\boldsymbol{E}_p$ as
\begin{equation}\label{eqn:normalize}
\boldsymbol{E}_p=\frac{\sum_{i\in \mathcal{N}} w_{p, i} \boldsymbol{e}_i+w_{\mathrm{bg}} \boldsymbol{e}_{\mathrm{bg}}}{\sum_{i\in \mathcal{N}} w_{p, i}+w_{\mathrm{bg}}},
\end{equation}
where $w_{\mathrm{bg}}$ and $\boldsymbol{e}_{\mathrm{bg}}$ are the weight and embedding feature of the background.
Because the weighted accumulation is commutative, the contribution of each Gaussian can be evaluated independently. We then adopt Eq.~\ref{eqn:LC-WSR} as the weight function $w_{p,i}$, which decreases with distance. Following LC-WSR, $\sigma$ is a learnable scene-scale parameter, $v_i$ is a view-dependent scalar decoded from additional SH coefficients, and $w_{\mathrm{bg}}$ is a learnable background weight; all are optimized jointly with the Gaussian parameters. We provide a detailed mathematical derivation of the blending formulation and its gradients in Appendix~\ref{app:oit_derivation}, showing that the pipeline remains end-to-end differentiable.

This reformulation expresses rendering as a sum of independently weighted embedding features. In the next section, we introduce how to reorganize the heavily FMA computation into batched matrix multiply-accumulate (MMA) operations executed on Tensor Cores.

\label{sec:OIFB}
\begin{figure}[t]
    \centering
    \includegraphics[width=\linewidth]{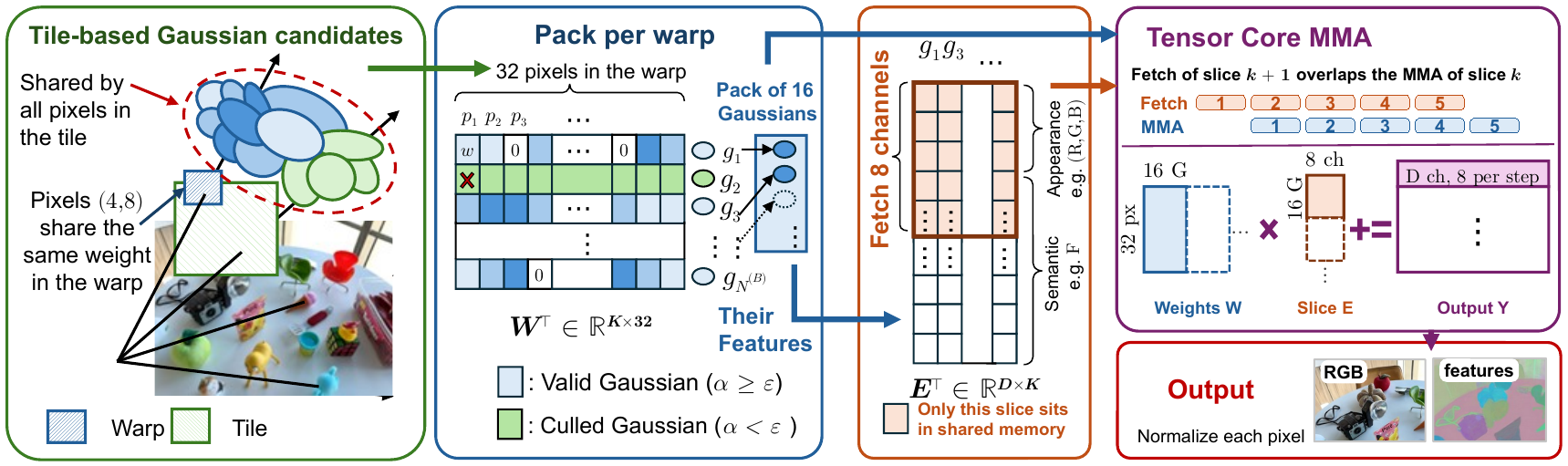}
    \vspace{-5mm}
    \caption{\textbf{Overview of our tile-based warp-level MMA rendering pipeline.} Each warp processes 32 pixels and packs 16 valid Gaussians into a weight matrix $\boldsymbol{W}$. Their features are fetched in 8-channel slices, so only the current slice resides in shared memory, and Tensor Cores multiply $\boldsymbol{W}$ with each slice while the next slice is being fetched.}%
    \label{fig:framework-new}

  \vspace{-2mm}
\end{figure}
\subsection{Packed Tile-Based Warp-Level MMA Execution}
\label{sec:MMA-execution}
The original 3DGS employs a tile-based rasterizer to achieve fast rendering. This method partitions the screen into $16 \times 16$ tiles where each tile is assigned to one thread block. All warps within the block collaboratively load the Gaussians overlapping the tile into shared memory. During tile-based rasterization, all warps in the same block share the same candidate Gaussian list,
\begin{equation}
\mathcal{G}^{(B)}=\{g_1,g_2,\cdots,g_{N^{(B)}}\},
\end{equation}
where $B$ denotes the block index and $N^{(B)}$ is the number of candidate Gaussians for the current tile.

To efficiently map semantic blending onto Tensor Core execution, the computation granularity should be aligned with the warp-level MMA primitive. Accordingly, we design a tile-based, warp-level accumulation pipeline. Specifically, we partition each $16 \times 16$ tile into multiple $4 \times 8$ pixel groups, each processed by a warp. For warp $u$ in block $B$, let $\mathcal{P}^{(u)}=\{p_1, p_2, \cdots, p_{32}\}$ denote the set of pixels handled by this warp, as shown on the left of Fig. \ref{fig:framework-new}.

Directly issuing MMA operations over all tile-level candidate Gaussians is inefficient because many Gaussians have negligible opacity for a specific pixel. During rasterization, Gaussian-pixel pairs whose projected opacity $\alpha$ is smaller than $\varepsilon$ are culled. For each candidate Gaussian $g$, we first compute a pixel-level validity mask and use it to obtain the masked rendering weight $\bar{w}_{p,g}^{(u)}$:
\begin{equation}
m_{p,g}^{(u)}
=
\boldsymbol{1}\left(\alpha_{p,g} \ge \varepsilon\right),
\qquad
\bar{w}_{p,g}^{(u)}
=
m_{p,g}^{(u)} w_{p,g},
\quad
p \in \mathcal{P}^{(u)}.
\end{equation}

If $\sum_{p \in \mathcal{P}^{(u)}} m_{p,g}^{(u)} = 0$, the Gaussian $g$ does not contribute to any of the 32 pixels in warp $u$ and is skipped for this warp-level MMA execution. Otherwise, it is packed into the current flush batch. This design is effective in practice because the zero-voting only performs within a single warp and the pixels with the same warp are more spatially adjacent and usually have similar valid Gaussians.

Once collecting $K_{\mathrm{mma}}=16$ valid Gaussians for the current warp, we form a packed weight matrix and a packed embedding features matrix:
\begin{equation}
\boldsymbol{W}^{(u,t)}
=
\left[
\bar{w}^{(u)}_{p_i,g_j}
\right]_{i=1,j=1}^{32,16}
\in \mathbb{R}^{32\times16},
\qquad
\boldsymbol{E}^{(u,t)}
=
\left[\boldsymbol{e}_{g_1}^{\top}, \boldsymbol{e}_{g_2}^{\top}, \cdots, \boldsymbol{e}_{g_{16}}^{\top}\right]^{\top}
\in \mathbb{R}^{16\times D}.
\end{equation}
Here, $t$ denotes the MMA flush-batch index, and $K_{\mathrm{mma}}=16$ is the reduction dimension of the \texttt{mma.m16n8k16} primitive used in our implementation, whose operands are stored in FP16 (Appendix~\ref{app:mma-shape}).
The packed batch is then accumulated through Tensor Core:
\begin{equation}
\boldsymbol{Y}^{(u)}
=
\sum_{t=0}^{\left\lceil N^{(u)} / K_{\mathrm{mma}} \right\rceil - 1}
\boldsymbol{W}^{(u,t)}
\boldsymbol{E}^{(u,t)} ,
\end{equation}
where $N^{(u)}$ denotes the number of Gaussians packed for warp $u$ after validity skip. Thus $\boldsymbol{W}^{(u)}\in\mathbb{R}^{32\times K}$ (drawn transposed in Fig.~\ref{fig:framework-new}) holds any number $K=N^{(u)}$ of packed Gaussians, and each MMA flush consumes 16 of them. The output $\boldsymbol{Y}^{(u)} \in \mathbb{R}^{32 \times D}$ holds the accumulated embeddings of the warp's pixels before the per-pixel normalization of Eq.~\ref{eqn:normalize}.

While the scheduler issues MMA operations for the current packed batch, the warp can continue buffering valid Gaussians from shared memory to registers, overlapping data preparation with MMA execution latency so that Tensor Cores stay busy across flush batches.
\subsection{Scalable MMA Rendering for High-Dimensional Embedded Features}

In the traditional render implementation, the whole feature vector of each overlapping Gaussian is preloaded into shared memory before blending. As the embedded feature dimension $D$ increases, the shared memory used by each block grows proportionally. This sharply reduces the number of resident warps on each SM and limits the ability of the scheduler to hide latency.

To address this issue, we adopt a lazy batched channel feature fetching strategy. We divide the feature dimension into a channel batch of 8, which is aligned with the MMA output-channel tile. With $C_{\mathrm{mma}}=8$, each MMA step computes an $8$-channel slice:
\begin{equation}
\boldsymbol{Y}^{(u)}_{:,\,8q:8q+7}
=
\sum_{t=0}^{\left\lceil N^{(u)} / K_{\mathrm{mma}} \right\rceil - 1}
\boldsymbol{W}^{(u,t)}
\boldsymbol{E}^{(u,t)}_{:,\,8q:8q+7},
\end{equation}
where $q=0,\cdots,\lceil D/8\rceil-1$ indexes the channel tile. At each flush MMA operation, the kernel loads only the required 8-dimensional feature slice for the current Gaussians, and the next slice is loaded while Tensor Cores process the current one (Fig.~\ref{fig:framework-new}). This design reduces shared-memory pressure while keeping the same matrix-oriented structure as low-dimensional rendering.

\section{Performance Evaluation}
\subsection{Experimental Setups}
\label{sec:setup}

\paragraph{Datasets \& Metrics.}
We evaluate our method on multiple datasets to assess both visual reconstruction quality and semantic feature performance. For image reconstruction, we use the Mip-NeRF 360 dataset~\cite{barron2021mip}, with segmentation labels annotated by GAGS~\cite{peng2026gags}, which provides complex visual scenes with associated semantic information. We report reconstruction fidelity using  Peak Signal-to-Noise Ratio (PSNR), Structural Similarity (SSIM), and Learned Perceptual Image Patch Similarity (LPIPS). To evaluate semantic feature quality, we follow LangSplat~\cite{qin2024langsplat} and conduct experiments on the LERF~\cite{kerr2023lerf} and 3D-OVS~\cite{liu2023weakly} benchmarks, which are specifically designed for language-driven 3D understanding tasks. On these datasets, we report mean Intersection over Union (mIoU) and localization accuracy for 3D semantic segmentation and object localization.

\paragraph{Baselines.}
We compare with 3DGS~\cite{kerbl20233d} using the LangSplat feature pipeline~\cite{qin2024langsplat} and with TC-GS~\cite{liao2025tc}. For order-independent methods, Sort-Free Gaussian Splatting~\cite{hou2024sort} has no official implementation, so we use a publicly available third-party implementation, and we retrain Duplex-GS~\cite{liu2026duplex} with its official code. For fast renderers, we use the official code of FlashGS~\cite{feng2024flashgs}, GEMM-GS~\cite{li2026gemmgs}, Speedy-Splat~\cite{hanson2025speedy}, and AdR-Gaussian~\cite{wang2024adrgaussian}, and integrate our method by replacing only their blending kernel while keeping their culling, scheduling, and sorting unchanged.

\paragraph{Implementation Details.}
To ensure fair comparison, we conduct all main experiments on an NVIDIA RTX 4090 GPU using the same settings as the original 3DGS~\cite{kerbl20233d}. For FPS evaluation, we perform several warm-up iterations prior to measurement and report results based on repeated timed runs with GPU synchronization before and after each render. Following TC-GS~\cite{liao2025tc}, we adopt a mixed-precision strategy that combines FP16 MMA operations with standard FP32 FMA operations. Main results use an end-to-end timing protocol, and comparisons with acceleration methods use the kernel-isolated protocol of FlashGS~\cite{feng2024flashgs} (Appendix~\ref{app:timing}). Results on RTX 6000 Ada and H100 GPUs, which show consistent speedups across architectures, are reported in Appendix~\ref{app:cross-gpu}.

\subsection{Quantitative Results}
\label{sec:quantitative}
\begin{table*}[t]
\centering
\small
\caption{\textbf{Quantitative comparison on Mip-NeRF 360 (OVS) and LERF.} We report image reconstruction quality, semantic performance, and end-to-end rendering speed. Following LangSplat~\cite{qin2024langsplat}, mIoU and localization accuracy (LocAcc) are evaluated based on rendered language features. FPS follows the end-to-end protocol (Appendix~\ref{app:timing}) on an RTX 4090. We use 3DGS equipped with the TC-GS \cite{liao2025tc} module as 3DGS(+TC-GS). Best results for each metric are \textbf{bold}; LPIPS is lower-is-better, and all other metrics are higher-is-better.}
\label{tab:mean-table}
\vspace{-2mm}
\setlength{\tabcolsep}{4pt}
\resizebox{\textwidth}{!}{%
\begin{tabular}{lcccccccccccc}
\toprule
& \multicolumn{6}{c}{Mip-NeRF 360} & \multicolumn{6}{c}{LERF} \\
\cmidrule(lr){2-7} \cmidrule(lr){8-13}
Method & PSNR & SSIM & LPIPS & mIoU & LocAcc & FPS & PSNR & SSIM & LPIPS & mIoU & LocAcc & FPS \\
\midrule
3DGS~\cite{kerbl20233d,qin2024langsplat} & 30.12 & 0.910 & 0.181 & 0.536 & 0.698 & 156.0 & 23.91 & \textbf{0.856} & 0.219 & 0.419 & 0.632 & 237.9 \\
3DGS(+TC-GS)~\cite{liao2025tc} & 30.14 & \textbf{0.911} & 0.183 & 0.536 & 0.698 & 295.1 & \textbf{23.97} & \textbf{0.856} & \textbf{0.218} & 0.419 & 0.632 & 310.3 \\
\midrule
MatrixSplat & \textbf{30.35} & \textbf{0.911} & \textbf{0.178} & \textbf{0.541} & \textbf{0.730} & \textbf{599.9} & 23.80 & 0.851 & 0.225 & \textbf{0.445} & \textbf{0.679} & \textbf{702.6} \\
\bottomrule
\end{tabular}%
}
\vspace{-3mm}
\end{table*}

Tab.~\ref{tab:mean-table} and Tab.~\ref{tab:mean-table-3DOVS} summarize the main quantitative results on Mip-NeRF 360~\cite{barron2021mip}, LERF~\cite{kerr2023lerf}, and 3D-OVS~\cite{liu2023weakly}. Following LangSplat \cite{qin2024langsplat}, mIoU and localization accuracy (LocAcc) are evaluated based on rendered language features. TC-GS's~\cite{liao2025tc} render quality remains very close to 3DGS because it does not change the rendering pipeline. Our method consistently achieves notable rendering speed improvements across all datasets while maintaining competitive reconstruction quality and semantic accuracy compared with both the vanilla 3DGS~\cite{kerbl20233d} and TC-GS~\cite{liao2025tc}. %

\begin{wraptable}{r}{0.6\linewidth}
\vspace{-4mm}
    \centering
    \captionof{table}{\textbf{Quantitative comparison on 3D-OVS.} Same setting as Tab.~\ref{tab:mean-table}.}%
    \vspace{-2mm}
    \label{tab:mean-table-3DOVS}
    \setlength{\tabcolsep}{5pt}
    \resizebox{\linewidth}{!}{%
    \begin{tabular}{lcccccc}
    \toprule
    & \multicolumn{6}{c}{3D-OVS} \\
    \cmidrule(lr){2-7}
    Method & PSNR & SSIM & LPIPS & mIoU & LocAcc & FPS \\
    \midrule
    3DGS~\cite{kerbl20233d,qin2024langsplat} & 24.71 & \textbf{0.757} & \textbf{0.234} & 0.802 & 0.937 & 162.0 \\
    3DGS(+TC-GS)~\cite{liao2025tc} & 24.66 & 0.749 & 0.234 & 0.802 & 0.937 & 294.8 \\
    \midrule
    MatrixSplat & \textbf{24.89} & 0.752 & 0.239 & \textbf{0.804} & \textbf{0.985} & \textbf{510.9} \\
    \bottomrule
    \end{tabular}%
    }
    \vspace{-1mm}
\end{wraptable}
On the Mip-NeRF 360 dataset, our method achieves 599.9 FPS, substantially outperforming the TC-GS by 2$\times$. Despite this large speedup, it maintains high visual quality and even achieves the best PSNR and LPIPS among all methods. On the LERF dataset, our method reaches 702.6 FPS, nearly 3$\times$ faster than 3DGS. Tab.~\ref{tab:mean-table-3DOVS} shows the quantitative comparison on 3D-OVS. Per-scene analysis shows that the fidelity change is scene-dependent: PSNR drops by 0.23\,dB on \textit{teatime}, where heavy semi-transparent overlap makes exact ordering important, but rises by 0.73\,dB on \textit{counter}, where avoiding view-dependent sorting errors helps. Nevertheless, the overall results show that our method delivers significant speedups while largely maintaining the quality of rendered features, including both appearance and language embeddings across all three datasets. %

\begin{wrapfigure}{r}{0.63\linewidth}
\vspace{-4mm}
\includegraphics[width=\linewidth]{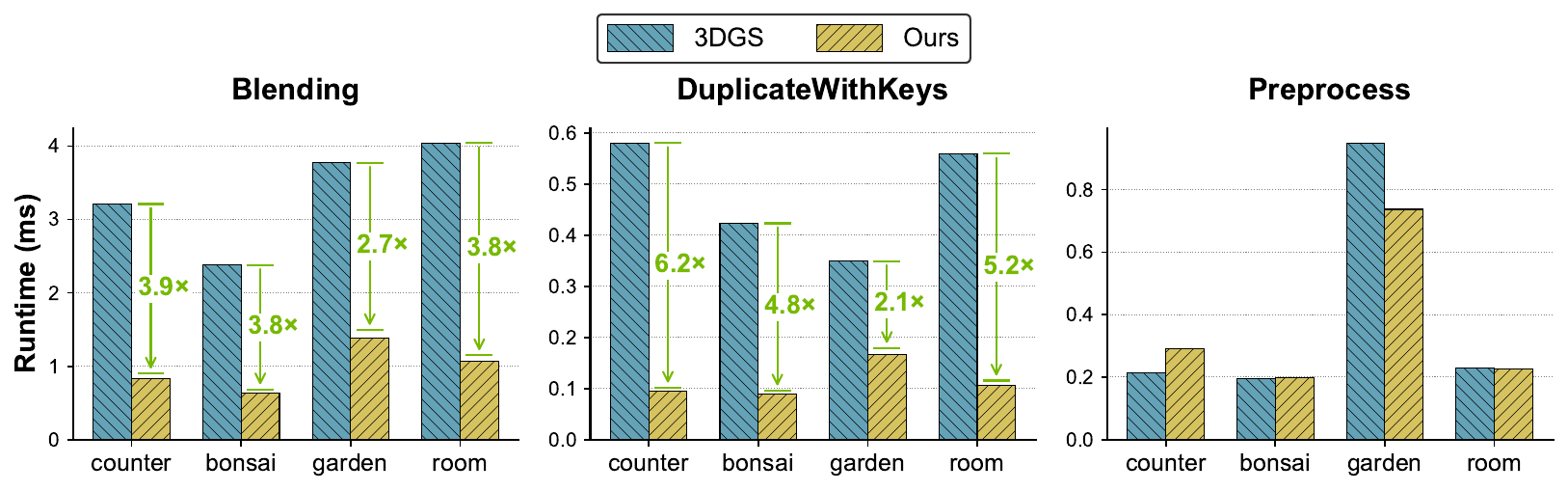}
    \vspace{-6mm}
    \caption{\textbf{Runtime breakdown of the rendering kernel on the Mip-NeRF 360 datasets.}}%
    \label{fig:kernel-stages}
    \vspace{-2mm}
\end{wrapfigure}
To further evaluate the improvement of our method for each stage, we conduct a comprehensive runtime breakdown of the most expensive kernels as listed in Sec.~\ref{sec:revisiting}: blending, DuplicateWithKeys, and preprocessing. Here, our blending kernel handles both rendering and sorting, so we combine the sorting and rendering runtime for 3DGS. Fig.~\ref{fig:kernel-stages} shows that our speedup mainly comes from the blending stages, which are also the dominant bottlenecks in the original 3DGS pipeline. Our Tensor-Core-oriented MMA computation reduces the latency of the render kernel by 2.1--3.0$\times$. In the 3DGS pipeline, each Gaussian-tile pair is associated with a depth-aware 64-bit key, making key construction and radix sorting expensive. In contrast, our method uses only the tile ID as key and completely removes depth from the sorting, eliminating the need for per-tile depth ordering and reducing the cost of key construction and radix sorting.

\subsection{Comparison with Order-Independent Methods and Fast Renderers}
\label{sec:comparison}
\begin{table*}[t]
\centering
\small
\caption{\textbf{Comparison with order-independent methods.} Kernel-isolated protocol (Appendix~\ref{app:timing}), $D{=}8$, RTX 4090. Speedups over 3DGS are in parentheses. Sort-Free uses a third-party implementation. $^*$Duplex-GS on 3D-OVS uses the official pretrained checkpoint; other Duplex-GS entries are retrained with the official code and default settings.}
\label{tab:comparison}
\vspace{-2mm}
\setlength{\tabcolsep}{4pt}
\resizebox{\textwidth}{!}{%
\begin{tabular}{lccccccccc}
\toprule
& \multicolumn{3}{c}{3D-OVS} & \multicolumn{3}{c}{LERF} & \multicolumn{3}{c}{Mip-NeRF 360} \\
\cmidrule(lr){2-4} \cmidrule(lr){5-7} \cmidrule(lr){8-10}
Method & FPS & PSNR & LPIPS & FPS & PSNR & LPIPS & FPS & PSNR & LPIPS \\
\midrule
3DGS~\cite{kerbl20233d} & 210.2 & 24.78 & 0.239 & 357.0 & 23.91 & \textbf{0.217} & 278.9 & 30.13 & 0.183 \\
Sort-Free (WSR)~\cite{hou2024sort} & 240.3 & 24.55 & 0.239 & 256.1 & 23.90 & 0.227 & 215.7 & 30.35 & \textbf{0.178} \\
Duplex-GS~\cite{liu2026duplex} & 319.1 & \textbf{25.40}$^*$ & \textbf{0.171}$^*$ & 332.9 & \textbf{24.46} & 0.223 & 127.9 & \textbf{30.46} & 0.222 \\
MatrixSplat & \textbf{521.8} (2.48$\times$) & 24.88 & 0.234 & \textbf{891.9} (2.50$\times$) & 23.80 & 0.228 & \textbf{660.0} (2.37$\times$) & 30.35 & 0.181 \\
\bottomrule
\end{tabular}%
}
\vspace{-3mm}
\end{table*}

\paragraph{Order-independent methods.}
Tab.~\ref{tab:comparison} separates the formulation from the execution strategy. Sort-Free rendering uses the same weighted-sum formulation with scalar execution and gains little speed, even running slower than 3DGS on LERF and Mip-NeRF 360. Duplex-GS attains the best PSNR, but its MLP evaluation adds overhead. Our method is 2.2--3.5$\times$ faster than Sort-Free at comparable PSNR and 1.6--5.2$\times$ faster than Duplex-GS, so the speedup comes from changing the compute primitive rather than from the formulation alone.

\paragraph{Integration with fast renderers.}
Pipeline-level accelerators reduce the work around blending but keep serial scalar accumulation. Replacing only their blending kernel with ours gives an additional 1.69--2.84$\times$ speedup across FlashGS, GEMM-GS, Speedy-Splat, and AdR-Gaussian with PSNR changes of at most 0.24\,dB, showing that the two kinds of optimization are complementary (Appendix~\ref{app:integration}). Our speedup over 3DGS is also stable across GPU architectures: 2.37$\times$, 2.36$\times$, and 2.33$\times$ on RTX 4090, RTX 6000 Ada, and H100, respectively (Appendix~\ref{app:cross-gpu}).

\subsection{Scalability with Feature Dimensionality}

\begin{wrapfigure}{r}{0.4\linewidth}
  \centering
  \vspace{-5mm}
  \includegraphics[width=\linewidth]{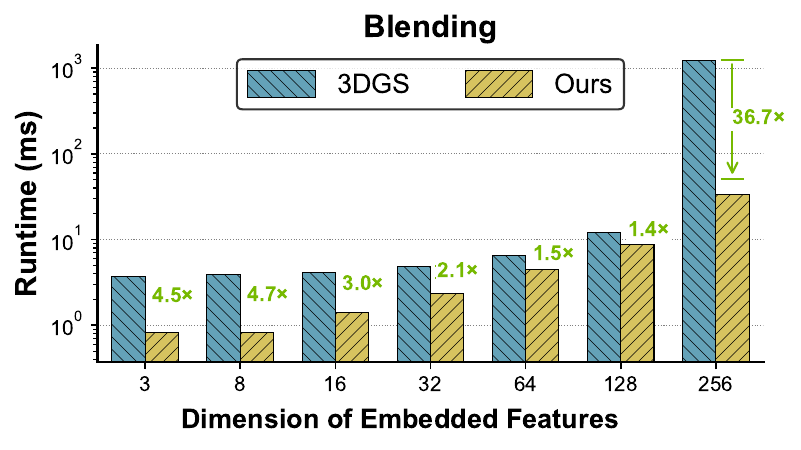}
  \vspace{-6mm}
  \caption{\textbf{Embedded feature scalability.} Our method consistently reduces blending kernel time over 3DGS as the feature dimension scales.} %
  \label{fig:sweep_plot}
  \vspace{-3mm}
\end{wrapfigure}
Fig.~\ref{fig:sweep_plot} shows that the blending kernel runtime of the original 3DGS renderer increases rapidly as the semantic feature dimension grows. Our method consistently outperforms 3DGS across all evaluated feature dimensions, reducing blending-kernel time by about 3$\times$ on average at low dimensions and by up to 36.7$\times$ at 256 dimensions. The large gain at high feature dimensions validates our lazy batched-channel feature fetching strategy. Instead of loading the full feature vector of each Gaussian into shared memory, we load only the 8-channel slice needed by the current MMA flush, so the shared-memory footprint and warp occupancy stay constant as $D$ grows. Our method maintains high efficiency even for high-dimensional semantic features, leading to better scalability for the render kernel.
\subsection{Visual Quality Comparisons}
\begin{figure}[t]
    \centering
    \includegraphics[width=\linewidth]{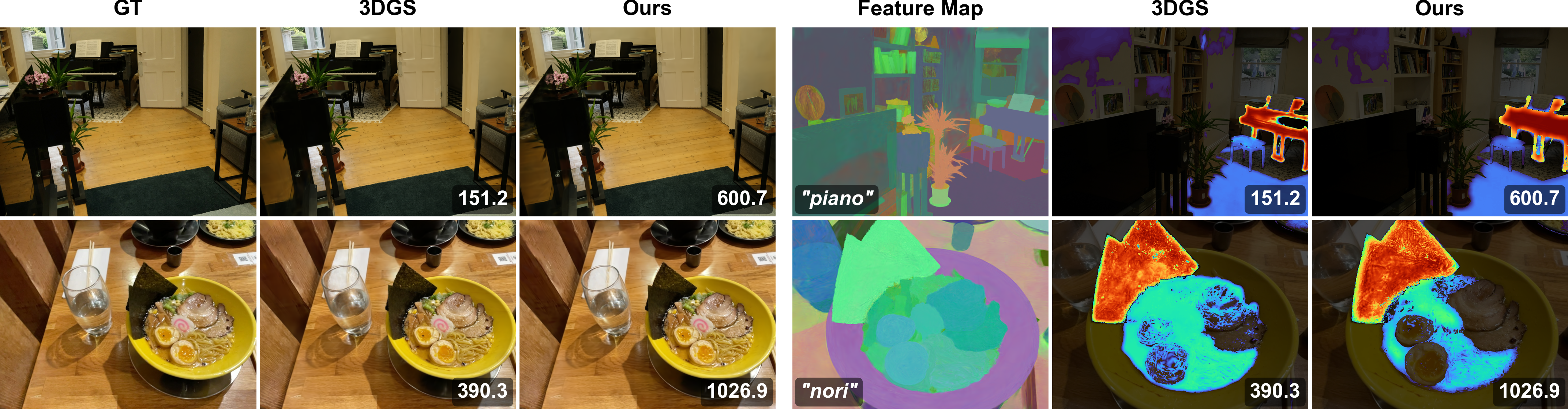}
    \vspace{-1mm}
    \caption{\textbf{Visual comparison of quality and language accuracy} on Room and Ramen. The number at the bottom right of each image is the rendering FPS.}
    \label{fig:comparison2}

  \vspace{-2mm}
\end{figure}

Fig.~\ref{fig:comparison2} compares 3DGS and our method on the room and ramen scenes. Our method better preserves the transparent water glass and the ramen texture while rendering at over 1,026 FPS, and it localizes queried regions such as the piano and nori with similar accuracy and sharpness. Additional comparisons are provided in Appendix~\ref{app:visual}. %

\subsection{Ablation Study}
\begin{table}[t]
  \centering
  \small
  \caption{\textbf{Component ablation across feature dimensions $D$.} Average FPS over 3D-OVS and Mip-NeRF 360 scenes at a lower resolution than Tab.~\ref{tab:comparison}; each row adds one component.}
  \label{tab:ablation}
  \vspace{1mm}
  \begin{tabular}{lrrrrrrr}
  \toprule
  Setting & $D{=}8$ & 16 & 32 & 64 & 128 & 256 & 512 \\
  \midrule
  Vanilla 3DGS                       & 253.8 & 237.2 & 194.1 & 116.3 & 59.5 & 2.3 & 1.1 \\
  + OIEFB                            & 423.5 & 363.2 & 273.5 & 143.7 & 64.9 & 2.1 & 1.2 \\
  + MMA execution                    & 741.1 & 500.9 & 302.8 & 148.1 & 78.8 & 38.0 & 17.4 \\
  + Warp-level packing               & 830.5 & 568.2 & 369.0 & 170.8 & 79.3 & 35.7 & 22.4 \\
  + Lazy channel-batched fetching    & \textbf{830.8} & \textbf{622.6} & \textbf{399.5} & \textbf{269.1} & \textbf{144.5} & \textbf{72.1} & \textbf{41.1} \\
  \bottomrule
  \end{tabular}
  \vspace{-3mm}
\end{table}
Tab.~\ref{tab:ablation} adds our components one at a time, averaged over all 3D-OVS and Mip-NeRF 360 scenes. Order-independent embedded feature blending (OIEFB, Sec.~\ref{sec:OIFB}) removes sorting and gives up to 1.67$\times$ at $D{=}8$, but it still uses scalar FMAs, so its benefit shrinks as $D$ grows. MMA execution is the main source of speedup. Warp-level packing reduces wasted MMA lanes and adds 12--22\% for $D\le64$. Lazy channel-batched fetching matters most at high dimensions, adding 1.6--2.0$\times$ for $D\ge64$. Overall, our method is 2.1--3.3$\times$ faster than 3DGS for $D\le128$ and 31--37$\times$ faster for $D\ge256$. Reduced precision alone does not explain the gain: running 3DGS with FP16 features changes its speed by at most 17\%, while our kernel is 2.2--4.8$\times$ faster on the same scenes; the gain instead comes from matrix execution and data reuse (Appendix~\ref{app:fp16}).

\begin{wrapfigure}{r}{0.32\linewidth}
    \centering
    \vspace{-5mm}
    \includegraphics[width=\linewidth]{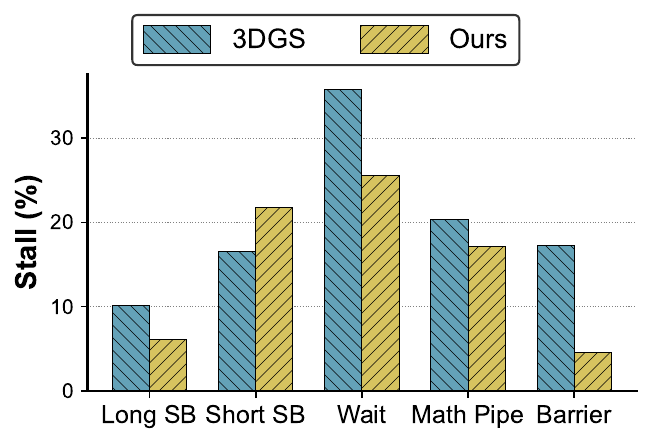}
    \vspace{-7mm}
    \caption{\textbf{Stall breakdown.}}
    \label{fig:stall-comparison}

  \vspace{-3mm}
\end{wrapfigure}
To demonstrate that order-independent blending alleviates the instruction dependencies introduced by alpha blending, we report the stall breakdown in Fig.~\ref{fig:stall-comparison} on 3DGS and our method. Our method reduces wait stalls by about 10\% by replacing the serial accumulation in alpha blending with order-independent accumulation. Since all lanes within a warp cooperatively execute the same MMA operation, our method reduces synchronization overhead. Therefore, barrier stall also drops about 10\%. Meanwhile, the MMA-based pipeline requires additional shared-memory staging for blending weights and intermediate matrices, which leads to increased Short Scoreboard stalls.

\section{Conclusion}
\paragraph{Summary.} We propose MatrixSplat, a matrix-parallel framework for feature-scalable 3D Gaussian Splatting rendering. We reformulate semantic feature rendering as an order-independent weighted accumulation and map it to warp-level packed MMA execution. We further introduce a lazy channel-batched feature-fetching strategy to reduce shared-memory pressure and maintain efficiency for high-dimensional features. Experiments on Mip-NeRF 360, LERF, and 3D-OVS show that MatrixSplat achieves up to 3.8$\times$ end-to-end speedup over 3DGS and 2$\times$ over TC-GS while largely maintaining visual quality and semantic accuracy, reduces render-kernel time by up to 36.7$\times$ for high-dimensional features, and provides an additional 1.69--2.84$\times$ speedup when integrated into existing fast renderers.
\paragraph{Limitations and Broader Impact.} First, order-independent accumulation approximates depth-ordered compositing, so its effect on fidelity is scene-dependent: it degrades quality under extreme multi-layer semi-transparent overlap (e.g., $-$0.23\,dB on \textit{teatime}) and can improve it under moderate occlusion, and thresholds tuned for alpha compositing may need re-tuning. Second, we accelerate forward rendering; training also becomes faster, but by less, because the backward pass still relies on scattered gradient updates and atomic operations; register-interleaved gradient accumulation with warp-level reduction is a promising next step. Third, tile binning with radix sort, per-Gaussian preprocessing, and the warp-level packing scan remain non-matrix stages, which bound the end-to-end speedup once blending is accelerated. Finally, we target NVIDIA Tensor Cores via warp-level MMA; other matrix units such as NPUs are not evaluated. We will release our CUDA implementation and evaluation scripts. For broader impact, our method enables more efficient feature rendering for applications such as interactive scene understanding~\cite{yu2026elsa3d,li2026toward}, language-guided 3D editing and generation~\cite{chi2025enhancing,chi2026balancing}, robotics~\cite{liu2026palm}, AR/VR, and 3D medical imaging~\cite{tang2026ct}.

\clearpage

\bibliographystyle{plainnat}

\bibliography{main}

\clearpage
\appendix

\section{Derivation of Order-Independent Embedded Features Blending}
\label{app:oit_derivation}

We provide the detailed mathematical formulation and gradient derivation for the weighted order-independent embedded features blending used in Sec.~\ref{sec:OIFB}.

\paragraph{Forward Rendering.}
Let $\mathcal{N}$ be the set of Gaussians overlapping a pixel $\boldsymbol{p}$, consistent with Eq.~\ref{eqn:render-color}.
Rasterizing Eq.~\ref{eqn:opacity-contribution} onto the image plane via EWA splatting projects the 3D covariance $\boldsymbol{\Sigma}_i$ to a 2D screen-space covariance $\boldsymbol{\Sigma}_{2D,i} = \boldsymbol{J}\boldsymbol{T}\boldsymbol{\Sigma}_i\boldsymbol{T}^\top\boldsymbol{J}^\top$, where $\boldsymbol{T}$ is the upper-left $3{\times}3$ submatrix of the view matrix and $\boldsymbol{J}$ is the Jacobian of the local affine approximation of the projective transformation. This yields the per-pixel spatial response
\begin{equation}
    G_i(\boldsymbol{p}) = \exp\!\left(-\tfrac{1}{2}\,\Delta \boldsymbol{p}_i^\top \boldsymbol{\Sigma}_{2D,i}^{-1}\,\Delta \boldsymbol{p}_i\right),
    \quad \Delta \boldsymbol{p}_i = \boldsymbol{p}- \boldsymbol{\mu}_i^{2D},
\end{equation}
where $\boldsymbol{\mu}_i^{2D}$ is the projected 2D center of Gaussian $i$. The projected (vanilla) alpha $\alpha_i = o_i \cdot G_i(\boldsymbol{p})$ is the 2D rasterization counterpart of Eq.~\ref{eqn:opacity-contribution}, with $o_i \in (0,1)$ the raw opacity.
Following previous work~\cite{hou2024sort} and Eq.~\ref{eqn:LC-WSR}, the effective weight is
\begin{equation}
    w_i =w(\alpha_i, d_i) = \alpha_i \max\!\left(0,\,1 - \frac{d_i}{\sigma}\right) v_i,
\end{equation}
where $v_i$ is the view-dependent visibility scalar evaluated from an additional spherical harmonics (SH), $d_i$ is the view-space depth of Gaussian $i$, and $\sigma$ is a learnable scene-scale parameter. Gaussians with $d_i \geq \sigma$ receive zero weight and are excluded from rendering.
The normalization factor includes a learnable background weight $w_{\mathrm{bg}}$:
\begin{equation}
    Z = \sum_{j \in \mathcal{N}} w_j + w_{\mathrm{bg}}.
\end{equation}
The rendered pixel color is then
\begin{equation}
\label{eqn:pixel_color}
    \boldsymbol{C} = \frac{\displaystyle\sum_{i \in \mathcal{N}} \boldsymbol{c}_i\,w_i + \boldsymbol{c}_{\mathrm{bg}}\,w_{\mathrm{bg}}}{Z},
\end{equation}
where $\boldsymbol{c}_i$ is the view-dependent color of Gaussian $i$ decoded from its SH coefficients, consistent with Eq.~\ref{eqn:render-color}.

Let $L$ be the loss function. We derive the gradients with respect to $\boldsymbol{c}_i$, $w_i$, $\alpha_i$, $o_i$, $v_i$, $\sigma$, and $w_{\mathrm{bg}}$ using the chain rule.%

\paragraph{Gradient w.r.t.\ Color $\boldsymbol{c}_i$.}
The partial derivative of the pixel color with respect to Gaussian $i$'s color is its normalized weight:
\begin{equation}
    \frac{\partial L}{\partial \boldsymbol{c}_i} = \frac{\partial L}{\partial \boldsymbol{C}} \cdot \frac{w_i}{Z}.
\end{equation}

\paragraph{Gradient w.r.t.\ Effective Weight $w_i$.}
Applying the quotient rule to Eq.~\ref{eqn:pixel_color} and summing over color channels $ch$:
\begin{equation}
\label{eqn:grad_wi}
    \frac{\partial L}{\partial w_i} = \frac{1}{Z}\sum_{ch} \frac{\partial L}{\partial \boldsymbol{C}^{ch}} \cdot \bigl(\boldsymbol{c}_i^{ch} - \boldsymbol{C}^{ch}\bigr).
\end{equation}
This formulation ensures that Gaussians whose colors $\boldsymbol{c}_i$ differ significantly from the current pixel estimate $\boldsymbol{C}$ receive larger gradients to adjust their contribution.

\paragraph{Gradient w.r.t.\ Projected Alpha $\alpha_i$.}
Let $\delta_i = \boldsymbol{1}[d_i < \sigma]$ be the indicator that Gaussian $i$ is within the depth cutoff. Since $w_i = \alpha_i \max(0, 1-d_i/\sigma) v_i$, by the chain rule:
\begin{equation}
    \frac{\partial L}{\partial \alpha_i}
    = \frac{\partial L}{\partial w_i} \cdot \delta_i \left(1 - \frac{d_i}{\sigma}\right) v_i.
\end{equation}

\paragraph{Gradient w.r.t.\ Raw Opacity $o_i$.}
Since $\alpha_i = o_i \cdot G_i$, and neglecting the boundary case of the $0.99$ clamp:
\begin{equation}
    \frac{\partial L}{\partial o_i}
    = \frac{\partial L}{\partial \alpha_i} \cdot G_i
    = \frac{\partial L}{\partial w_i} \cdot \delta_i\, G_i \left(1 - \frac{d_i}{\sigma}\right) v_i.
\end{equation}

\paragraph{Gradient w.r.t.\ View-Dependent Scalar $v_i$.}
\begin{equation}
    \frac{\partial L}{\partial v_i}
    = \frac{\partial L}{\partial w_i} \cdot \delta_i\, \alpha_i \left(1 - \frac{d_i}{\sigma}\right).
\end{equation}
This gradient is backpropagated through the SH basis to update the additional SH coefficients $\{a_k^{(i)}\}$:
\begin{equation}
    \frac{\partial L}{\partial a_k^{(i)}} = \frac{\partial L}{\partial v_i} \cdot Y_k(\hat{\boldsymbol{d}}_i),
\end{equation}
where $Y_k$ are the real spherical harmonic basis functions and $\hat{\boldsymbol{d}}_i$ is the unit view direction toward Gaussian $i$.

\paragraph{Gradient w.r.t.\ Scale Parameter $\sigma$.}
The gradient is accumulated over all Gaussians within the depth cutoff ($d_i < \sigma$):
\begin{equation}
    \frac{\partial L}{\partial \sigma}
    = \sum_{i \in \mathcal{N}} \frac{\partial L}{\partial w_i} \cdot \delta_i\, \alpha_i\, v_i \cdot \frac{d_i}{\sigma^2}.
\end{equation}

\paragraph{Gradient w.r.t.\ Background Weight $w_{\mathrm{bg}}$.}
Analogously to Eq.~\ref{eqn:grad_wi}:
\begin{equation}
    \frac{\partial L}{\partial w_{\mathrm{bg}}}
    = \frac{1}{Z}\sum_{ch} \frac{\partial L}{\partial \boldsymbol{C}^{ch}} \cdot \bigl(\boldsymbol{c}_{\mathrm{bg}}^{ch} - \boldsymbol{C}^{ch}\bigr).
\end{equation}

This formulation remains fully end-to-end differentiable, which changes the compositing model from depth-ordered alpha blending in Eq.~\ref{eqn:render-color} to an order-independent weighted approximation. We empirically study its impact on reconstruction and semantic rendering quality.

\section{Impact of MMA Shape}
\label{app:mma-shape}
  \begin{wraptable}{r}{0.42\linewidth}
  \centering
  \vspace{-4mm}
  \caption{\textbf{Rendering performance on two MMA shapes.}}
  \label{tab:ablation-batchsize}
  \begin{tabular}{c c c r}
  \toprule
   MMA Shape & FPS \\
  \midrule
   mma.m16n8k8      & 534.0 \\
   mma.m16n8k16   & 564.4 \\
  \bottomrule
  \end{tabular}
  \end{wraptable}
Tab.~\ref{tab:ablation-batchsize} reports the rendering performance of two MMA shapes: \texttt{mma.m16n8k8} and \texttt{mma.m16n8k16}. Compared with \texttt{mma.m16n8k8}, \texttt{mma.m16n8k16} processes twice the reduction dimension per MMA instruction, thereby reducing the number of K-tiles and flush operations. In our implementation, each flush stage starts with warp-level \texttt{ldmatrix} operations and a synchronization barrier. Therefore, \texttt{mma.m16n8k16} amortizes fixed pipeline overheads over more useful computation, leading to higher computational efficiency.

\section{Timing Protocols}
\label{app:timing}
We use two timing protocols. The \textit{end-to-end} protocol (Tabs.~\ref{tab:mean-table} and~\ref{tab:mean-table-3DOVS}) times the full per-frame \texttt{render()} call. The \textit{kernel-isolated} protocol (Tabs.~\ref{tab:comparison} and~\ref{tab:ablation} and the tables in this appendix) follows the measurement script of FlashGS~\cite{feng2024flashgs}, which is also used by Speedy-Splat~\cite{hanson2025speedy} and GEMM-GS~\cite{li2026gemmgs}. The two protocols differ in timing scope. The end-to-end protocol measures the full per-frame \texttt{render()} path, including per-frame activations, buffer setup, and rasterizer construction. The kernel-isolated protocol prepares the view-independent state once and times only the CUDA kernels; for the same checkpoint, GPU, and view, this changes the per-frame runtime from 1.32\,ms to 0.93\,ms. The end-to-end protocol reflects the cost an application experiences in practice, which is why we use it for the main results, while the kernel-isolated protocol allows a fair comparison with acceleration methods under their own convention. Absolute FPS therefore differs between the two protocols (e.g., 156.0 vs.\ 278.9 FPS for 3DGS on Mip-NeRF 360), while speedups within each table are directly comparable.

\section{Hardware Profiling: Precision versus Execution Structure}
\label{app:fp16}
Since our MMA kernel uses FP16 operands, we check whether its speedup comes from reduced precision alone by running the standard 3DGS blending with FP16 features, converted once outside the timed region.

\begin{table}[h]
  \centering
  \caption{\textbf{Nsight Compute counters} per launch ($D{=}32$, \textit{garden}).}
  \label{tab:ncu}
  \begin{tabular}{lrrr}
    \toprule
    Counter & Scalar FP32 & Scalar FP16 & Warp-packed MMA \\
    \midrule
    Executed warp instr. & 1341\,M & 1575\,M & \textbf{501\,M} ($-$63\%) \\
    FMA-pipe instr. & 630\,M & 763\,M & 171\,M \\
    ALU-pipe instr. & 217\,M & 217\,M & 175\,M \\
    Tensor-Core instr. & 0 & 0 & 4.17\,M \\
    DRAM bytes & 391\,MB & 319\,MB & 306\,MB \\
    L2 bytes & 844\,MB & 679\,MB & 745\,MB \\
    L1 bytes & 7.99\,GB & 7.99\,GB & \textbf{0.98\,GB} ($-$88\%) \\
    DRAM BW utilization & 14.5\% & 12.2\% & 30.5\% \\
    \bottomrule
  \end{tabular}
\end{table}

FP16 storage alone has little influence on 3DGS: on \textit{counter} and \textit{garden} with $D$ from 8 to 128, it changes throughput by $-$3\% to $+$17\%, whereas our kernel is 2.2--4.8$\times$ faster than FP32 3DGS. Tab.~\ref{tab:ncu} shows why: FP16 storage reduces DRAM and L2 traffic, but values are still converted inside the serial blending loop. Even a redesigned \texttt{half2} kernel reaches only 1.14$\times$, because the blending weights $w_i=\alpha_i T_i$ depend on $T_{i+1}=T_i(1-\alpha_i)$ and cannot be vectorized across Gaussians. Our order-independent formulation removes this recurrence, so the warp-packed MMA kernel executes 63\% fewer warp instructions and moves 88\% less L1 data.

\section{Integration with Fast Renderers}
\label{app:integration}
\begin{table}[h]
\centering
\small
\caption{\textbf{Integration into fast renderers.} Kernel-isolated protocol (Appendix~\ref{app:timing}), $D{=}8$, RTX 4090. Speedups in parentheses are relative to the host renderer.}
\label{tab:integration}
\setlength{\tabcolsep}{5pt}
\resizebox{\textwidth}{!}{%
\begin{tabular}{lcccccc}
\toprule
& \multicolumn{2}{c}{3D-OVS} & \multicolumn{2}{c}{LERF} & \multicolumn{2}{c}{Mip-NeRF 360} \\
\cmidrule(lr){2-3} \cmidrule(lr){4-5} \cmidrule(lr){6-7}
Method & FPS & PSNR & FPS & PSNR & FPS & PSNR \\
\midrule
FlashGS~\cite{feng2024flashgs} & 794.7 & 24.71 & 1432.2 & 23.99 & 996.4 & 30.06 \\
FlashGS + MatrixSplat & \textbf{2122.4} (2.67$\times$) & 24.81 & \textbf{3000.5} (2.09$\times$) & 23.81 & \textbf{2302.5} (2.31$\times$) & 30.30 \\
\midrule
GEMM-GS~\cite{li2026gemmgs} & 229.2 & 24.63 & 439.8 & 23.97 & 324.1 & 30.10 \\
GEMM-GS + MatrixSplat & \textbf{651.0} (2.84$\times$) & 24.79 & \textbf{1248.0} (2.84$\times$) & 23.80 & \textbf{911.4} (2.81$\times$) & 30.31 \\
\midrule
Speedy-Splat~\cite{hanson2025speedy} & 556.7 & 24.65 & 1000.2 & 23.96 & 704.6 & 30.13 \\
Speedy-Splat + MatrixSplat & \textbf{1252.0} (2.25$\times$) & 24.79 & \textbf{1774.3} (1.77$\times$) & 23.80 & \textbf{1360.3} (1.93$\times$) & 30.35 \\
\midrule
AdR-Gaussian~\cite{wang2024adrgaussian} & 489.1 & 24.64 & 952.9 & 23.97 & 654.6 & 30.09 \\
AdR-Gaussian + MatrixSplat & \textbf{944.0} (1.93$\times$) & 24.75 & \textbf{1607.4} (1.69$\times$) & 23.81 & \textbf{1201.4} (1.84$\times$) & 30.31 \\
\bottomrule
\end{tabular}%
}
\end{table}
Tab.~\ref{tab:integration} reports the four pipeline-level accelerators with and without our blending kernel. In each case we replace only the blending kernel and keep the host renderer's culling, scheduling, and sorting. The additional speedup ranges from 1.69$\times$ to 2.84$\times$, and PSNR changes by at most 0.24\,dB. FlashGS + MatrixSplat reaches 2302.5 FPS on Mip-NeRF 360.

\section{Cross-GPU Results}
\label{app:cross-gpu}
\begin{table}[h]
  \centering
  \caption{\textbf{Cross-GPU results.} FPS on Mip-NeRF 360 under the kernel-isolated protocol ($D{=}8$). Speedups in parentheses are relative to the baseline in the row above.}
  \label{tab:cross-gpu}
  \vspace{1mm}
  \setlength{\tabcolsep}{10pt}
  \begin{tabular}{lccc}
    \toprule
    Method & RTX 4090 & RTX 6000 Ada & H100 80GB \\
    \midrule
    3DGS & 278.9 & 252.1 & 252.5 \\
    3DGS + MatrixSplat & \textbf{660.0} (2.37$\times$) & \textbf{595.4} (2.36$\times$) & \textbf{587.5} (2.33$\times$) \\
    \midrule
    FlashGS & 996.4 & 932.7 & 769.7 \\
    FlashGS + MatrixSplat & \textbf{2302.5} (2.31$\times$) & \textbf{2190.3} (2.35$\times$) & \textbf{1904.7} (2.47$\times$) \\
    \bottomrule
  \end{tabular}
\end{table}
Tab.~\ref{tab:cross-gpu} reports results on consumer Ada (RTX 4090), workstation Ada (RTX 6000 Ada), and datacenter Hopper (H100) GPUs. The speedups are stable at 2.33--2.37$\times$ over 3DGS and 2.31--2.47$\times$ over FlashGS. Our implementation relies only on standard warp-level MMA instructions and shared memory, and does not depend on features specific to the RTX 4090.

\section{Additional Visual Quality Comparisons}
\label{app:visual}

\begin{figure}[t]
    \centering
    \includegraphics[width=\linewidth]{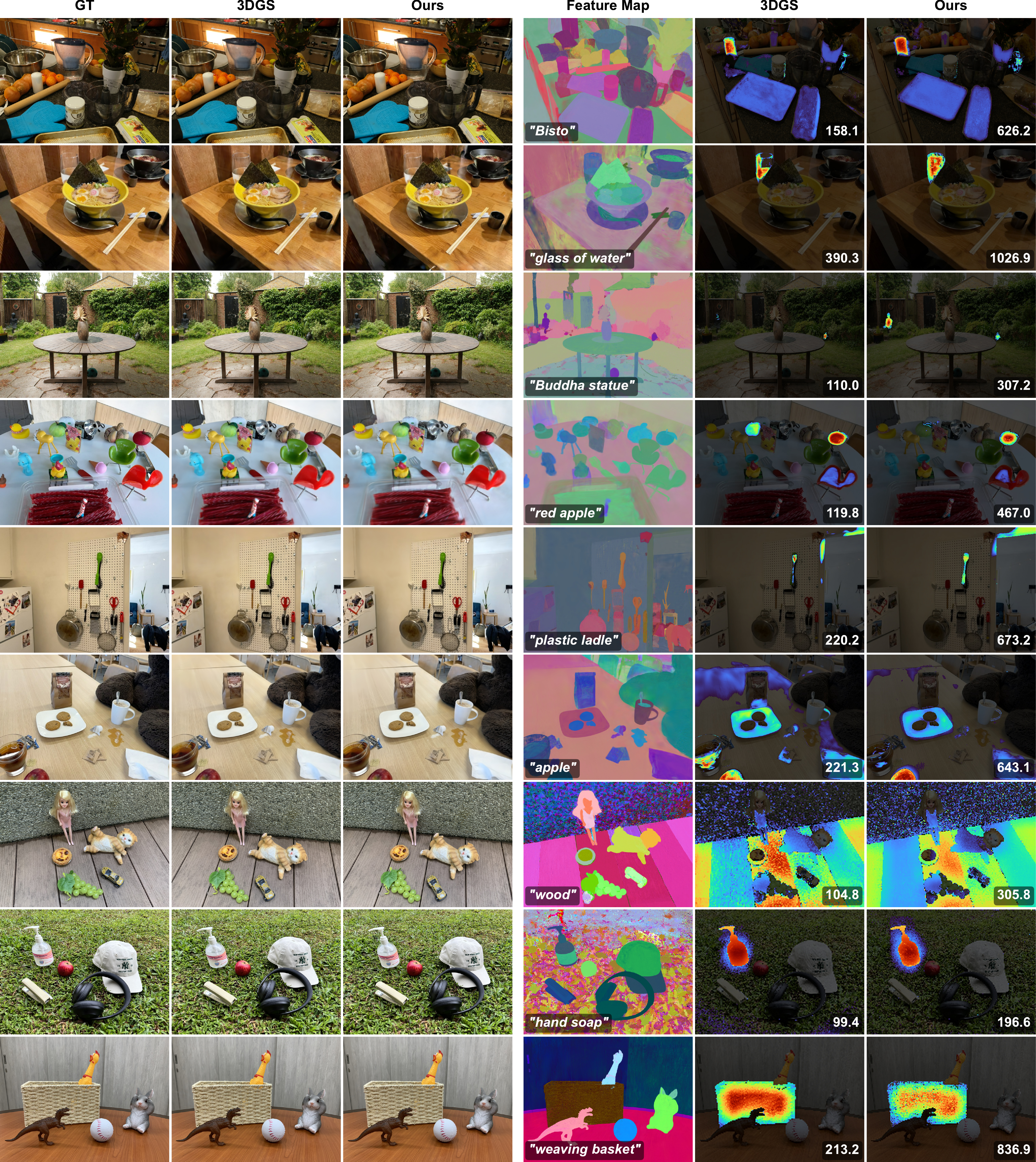}
    \vspace{-5mm}
    \caption{\textbf{Visual comparison of quality and language accuracy.} We compare the 3DGS and our method on varying datasets. The number shown at the bottom right is rendering FPS. Following LangSplat~\cite{qin2024langsplat}, the relevancy map is evaluated based on rendered language features.}
    \label{fig:comparison-appendix}

  \vspace{-2mm}
\end{figure}

We provide additional visual comparisons in Fig.~\ref{fig:comparison-appendix}. The number shown at the bottom right is rendering FPS. Following LangSplat~\cite{qin2024langsplat}, the relevancy map is evaluated based on rendered language features.
\clearpage

\end{document}

%% file: preamble.tex
\definecolor{miao}{RGB}{140,114,143}

\usepackage{graphicx}
\usepackage{algorithmicx}
\usepackage{algorithm}
\usepackage{algpseudocode}
\usepackage{wrapfig}
\usepackage[most]{tcolorbox}
\usepackage{bbm}

\newtcbox{\topone}{on line, colback=first, colframe=first, boxrule=0pt, arc=0pt, boxsep=1pt, left=2pt, right=2pt, top=1pt, bottom=1pt}
\newtcbox{\toptwo}{on line, colback=second, colframe=second, boxrule=0pt, arc=0pt, boxsep=1pt, left=2pt, right=2pt, top=1pt, bottom=1pt}
\newtcbox{\greenbar}{on line, colback=green, colframe=green, boxrule=0pt, arc=0pt, boxsep=1pt, left=2pt, right=2pt, top=1pt, bottom=1pt}

\usepackage{soul}
\usepackage{adjustbox}
\usepackage{bm}
\usepackage{algorithmicx}
\usepackage{algorithm}
\usepackage{algpseudocode}
\usepackage{pifont}
\usepackage{wasysym}
\usepackage{placeins} %
\usepackage{graphicx}
\usepackage{multirow}
\usepackage[hang,flushmargin]{footmisc}
\usepackage{balance}
\usepackage{latexsym}
\usepackage[most]{tcolorbox}
\usepackage{float}
\usepackage{booktabs}
\usepackage{placeins} %
\usepackage{multirow} %
\usepackage{url} %
\usepackage[misc]{ifsym}
\definecolor{first}{HTML}{FF9999}
\definecolor{second}{HTML}{FFCC99}
\definecolor{yellow}{HTML}{FFF8AD}
\definecolor{green}{HTML}{85D0FF}%
\usepackage{listings}

\usepackage{tikz}

\usepackage{framed}

\definecolor{that-yellow}{HTML}{ddc26f}

\makeatletter
\let\@authorsaddresses\@empty
\makeatother